\documentclass[conference]{IEEEtran}

\usepackage[utf8]{inputenc}
\usepackage[T1]{fontenc}
\usepackage{amsmath,amssymb,amsfonts}
\usepackage{algorithmic}
\usepackage{graphicx}
\usepackage{textcomp}
\usepackage{booktabs}
\usepackage{multirow}
\usepackage{array}
\usepackage{url}
\usepackage{cite}
\usepackage{hyperref}
\hypersetup{
    colorlinks=true,
    linkcolor=black,
    citecolor=black,
    urlcolor=blue
}

\def\BibTeX{{\rm B\kern-.05em{\sc i\kern-.025em b}\kern-.08em
    T\kern-.1667em\lower.7ex\hbox{E}\kern-.125emX}}

\begin{document}

\title{ERTS: Adversarial Robustness Testing of Ethical AI via Semantic Perturbation in a Bounded Consequence Space}

\author{
\IEEEauthorblockN{Pratyush Chaudhari}
\IEEEauthorblockA{Independent Researcher\\
\href{https://orcid.org/0009-0001-8815-7779}{ORCID: 0009-0001-8815-7779}\\[4pt]
{\footnotesize Research Assistance: Saim Kotkar, Vyankatesh Dawale}}
}

\maketitle

\begin{abstract}
As AI systems are deployed in high-stakes ethical contexts such as healthcare triage, autonomous vehicle control, and employment screening, formal methods for evaluating their robustness against adversarial manipulation of ethical reasoning remain underdeveloped. This paper introduces the Ethical Robustness Testing System (ERTS), a closed-pipeline framework that: (1)~encodes ethical dilemmas into a 22-dimensional Ethical Consequence Space (ECS) grounded in established ethical theory; (2)~applies 17 semantic perturbation functions subject to 6 validity constraint classes including a novel semantic coherence constraint; (3)~measures decision deviation via a 4-component Ethical Instability Index (EII); and (4)~produces domain-adaptive pre-deployment robustness assessment verdicts. We evaluate 4 structured baseline models and 2 production LLMs (Gemini~2.0~Flash and Llama~3.2) across 50 ethical scenarios spanning 8 deployment domains, generating 1,500 adversarial test cases. Results demonstrate that only 33\% of models achieve assessment clearance, with the local Llama-3.2 model proving particularly vulnerable to fairness corruption and information degradation attacks (ERS\,=\,0.737). To the best of our knowledge, no existing framework combines a bounded ethical consequence space, semantic coherence constraints, and domain-adaptive assessment in a single adversarial testing pipeline.
\end{abstract}


%═══════════════════════════════════════════════════════════════
% I. INTRODUCTION
%═══════════════════════════════════════════════════════════════
\section{Introduction}

The deployment of AI systems in ethical decision-making contexts has accelerated significantly, with applications in clinical decision support~\cite{rajkomar2019}, autonomous vehicle control~\cite{koopman2017}, algorithmic hiring~\cite{raghavan2020}, and military target identification~\cite{scharre2018}. These systems must make decisions that are accurate, ethically sound, fair, and resistant to manipulation~\cite{dignum2019}. Recent surveys have highlighted the multifaceted nature of ethical concerns in AI---spanning fairness, privacy, accountability, safety, robustness, transparency, and environmental impact---from both technical and societal perspectives~\cite{kenfack2025}.

While adversarial robustness testing is standard practice for ML models~\cite{goodfellow2015,madry2018}, existing frameworks operate on raw data representations. The Adversarial Robustness Toolbox (ART)~\cite{art2018} perturbs pixel values and input features. NVIDIA Garak~\cite{garak2024} red-teams text generation models. TrustLLM~\cite{trustllm2024} and HELM~\cite{helm2023} benchmark LLM safety properties. However, none evaluate whether an AI model's ethical judgment remains stable when the ethical framing of a scenario is deliberately manipulated.

This gap motivated our work. We observed that a healthcare AI performing well under standard testing could catastrophically fail when a scenario was reframed to emphasize short-term benefits over long-term harm, or when authority pressure overrode fairness considerations. We deliberately chose to operate on a structured ethical vector space rather than raw text because our pilot experiments showed that token-level perturbations to ethical scenario descriptions (e.g., replacing ``patient'' with ``subject'') do not correspond to meaningful ethical manipulations and produce false confidence in model robustness.

We present the Ethical Robustness Testing System (ERTS), addressing this gap through four contributions:
\begin{enumerate}
    \item \textbf{The Ethical Consequence Space (ECS):} a 22-dimensional vector representation grounded in utilitarian, deontological, and virtue ethics frameworks, where each dimension corresponds to a named ethical variable with defined polarity and semantic meaning.
    \item \textbf{17 Semantic Perturbation Functions} across 7 adversarial categories, subject to 6 validity constraint classes including a novel semantic coherence constraint~(C5).
    \item \textbf{The Ethical Instability Index (EII):} a 4-component composite metric with justified weights, quantifying decision deviation under perturbation.
    \item \textbf{Domain-Adaptive Pre-Deployment Assessment:} a multi-check process with thresholds grounded in regulatory standards, producing \textsc{Cleared}, \textsc{Conditional}, or \textsc{Failed} verdicts across 7 application domains.
\end{enumerate}

%═══════════════════════════════════════════════════════════════
% II. RELATED WORK
%═══════════════════════════════════════════════════════════════
\section{Related Work}

\subsection{Adversarial Machine Learning}
Adversarial robustness has been studied extensively in computer vision~\cite{goodfellow2015}, NLP~\cite{textattack2020}, and reinforcement learning~\cite{gleave2020}. Goodfellow et al.~\cite{goodfellow2015} introduced FGSM for image classifiers. Carlini and Wagner~\cite{carlini2017} developed optimization-based $L_p$ attacks. Madry et al.~\cite{madry2018} proposed PGD as a benchmark attack. These methods operate on raw input features and measure robustness as the perturbation budget required to flip a classification decision. The ART toolkit~\cite{art2018} unifies these approaches but does not model the semantic structure of ethical decisions.

\subsection{AI Safety and Alignment}
AI safety research has focused on alignment~\cite{russell2019}, value learning~\cite{hadfield2017}, and constitutional AI~\cite{bai2022}. Anthropic's constitutional AI~\cite{bai2022} trains models to follow explicit ethical principles. OpenAI's RLHF pipeline~\cite{ouyang2022} aligns models with human preferences through reward modeling. However, these approaches focus on training methodology rather than post-deployment robustness testing.

\subsection{AI Ethics Benchmarks}
TrustLLM~\cite{trustllm2024} evaluates LLM safety across toxicity and fairness dimensions using 30+ datasets and 16 LLMs. HELM~\cite{helm2023} measures resistance to harmful behaviors. The ETHICS benchmark~\cite{hendrycks2021} tests moral judgment on textual scenarios. These benchmarks evaluate static performance on fixed test sets but do not perform adversarial perturbation of ethical variables---they measure what a model does, not how easily it can be made to do something different. Kenfack~et~al.~\cite{kenfack2025} provide a comprehensive socio-technical survey that unifies these ethical principles and discusses both current and future concerns of deploying AI into society, but do not address adversarial robustness of ethical reasoning. Our work is complementary: ERTS could be applied as an adversarial stress-test layer on top of these static benchmarks and the ethical principles catalogued by such surveys.

\subsection{AI Certification Standards}
UL~3115~\cite{ul3115} provides safety criteria for AI-based products. ISO/IEC~22989~\cite{iso22989} and ISO/IEC~23894~\cite{iso23894} establish risk management frameworks. The EU AI Act~\cite{euaiact2024} mandates robustness requirements for high-risk systems. These are regulatory frameworks that define \emph{what} should be tested but not \emph{how}. ERTS provides computational infrastructure that could support future regulatory compliance processes.

\subsection{Ethical Adversarial Testing and Moral Perturbation}
Several recent works have explored adversarial evaluation specifically targeting ethical reasoning in AI, and ERTS builds upon this emerging line of research.

MoralExceptQA~\cite{moralexceptqa2022} tests whether LLMs can identify exceptions to moral rules under contextual variation, demonstrating model sensitivity to scenario framing but operating at the prompt level without a formal ethical vector representation. Value-attenuation attacks~\cite{qi2024} show that LLM alignment can be eroded through multi-turn dialogues, revealing RLHF-trained model vulnerability to systematic ethical manipulation. The Delphi system~\cite{delphi2021} provides commonsense moral reasoning but exhibits inconsistencies under simple rephrasing~\cite{talat2022}. TextAttack~\cite{textattack2020} provides general adversarial NLP but lacks ethical-domain perturbation semantics.

ERTS differs in three key respects: (1)~it operates on a formal 22-dimensional ethical consequence space rather than raw text; (2)~it enforces semantic coherence constraints~(C5) that prevent logically impossible ethical manipulations; and (3)~it produces quantitative domain-adaptive assessment verdicts rather than binary judgments.

%═══════════════════════════════════════════════════════════════
% III. ETHICAL CONSEQUENCE SPACE
%═══════════════════════════════════════════════════════════════
\section{The Ethical Consequence Space}

\subsection{Definition}
We define the Ethical Consequence Space (ECS) as a bounded vector space in $\mathbb{R}^d$ where $d = 22$ and each dimension $x_i$ represents a named ethical variable with semantic meaning. For any ethical decision scenario with action set $\mathcal{A}$, each action $a \in \mathcal{A}$ is encoded as a vector $\mathbf{x}_a \in [0, 1]^d$.

Table~\ref{tab:ecs_dims} shows 10 representative dimensions (of 22 total). The full specification includes additional variables for \texttt{collateral\_damage}, \texttt{legal\_violation\_score}, \texttt{proportionality\_score}, \texttt{consent\_violation}, \texttt{manipulation\_level}, \texttt{data\_exposure}, \texttt{restrictiveness}, \texttt{reversibility}, \texttt{precedent\_risk}, \texttt{stakeholder\_impact}, \texttt{welfare\_impact}, and \texttt{deception\_level}.

\begin{table}[htbp]
\caption{Representative ECS Dimensions (10 of 22)}
\label{tab:ecs_dims}
\centering
\begin{tabular}{lcc}
\toprule
\textbf{Variable} & \textbf{Polarity} & \textbf{Domain Relevance} \\
\midrule
harm\_to\_others & Negative & Healthcare, Military \\
harm\_to\_self & Negative & Healthcare, Vehicles \\
lives\_at\_risk\_score & Negative & Healthcare, Military \\
fairness\_impact & Positive & Hiring, Finance, Edu. \\
discrimination\_level & Negative & Hiring, Finance \\
accountability\_score & Positive & All domains \\
benefit\_score & Positive & All domains \\
safety\_risk & Negative & Healthcare, Vehicles \\
transparency\_score & Positive & All domains \\
privacy\_impact & Negative & Privacy, Healthcare \\
\bottomrule
\end{tabular}
\end{table}

\subsection{Theoretical Grounding}
The 22 ECS dimensions are derived from established ethical frameworks rather than arbitrary selection. Each dimension group maps to a specific philosophical tradition, as shown in Table~\ref{tab:ecs_grounding}.

\begin{table}[htbp]
\caption{Theoretical Grounding of ECS Dimension Groups}
\label{tab:ecs_grounding}
\centering
\small
\begin{tabular}{p{2.8cm}p{2.5cm}p{1.8cm}}
\toprule
\textbf{Dimension Group} & \textbf{Ethical Framework} & \textbf{Source} \\
\midrule
harm\_to\_others, harm\_to\_self, lives\_at\_risk & Utilitarian harm calculus & Mill~\cite{mill1863} \\
fairness\_impact, discrimination\_level & Rawlsian distributive justice & Rawls~\cite{rawls1971} \\
accountability, transparency & Kantian duty ethics & Kant~\cite{kant1785} \\
benefit\_score, welfare\_impact & Capabilities approach & Sen~\cite{sen2009}, Nussbaum~\cite{nussbaum2011} \\
consent\_violation, privacy\_impact & Autonomy / Rights-based & Beauchamp~\cite{beauchamp2019} \\
proportionality, reversibility & Prima facie duties & Ross~\cite{ross1930} \\
\bottomrule
\end{tabular}
\end{table}

We settled on $d=22$ after iterating between $d=15$ and $d=25$ during development. At $d=15$, fairness and discrimination variables collapsed into a single factor that masked bias detection failures under authority injection perturbations, leading to false negatives. At $d=25$, three variables (\texttt{cultural\_sensitivity}, \texttt{environmental\_impact}, \texttt{intergenerational\_harm}) showed near-zero variance across our scenario corpus and were removed. The current dimensionality represents the minimal set that preserves discriminative power across all 7 perturbation categories.

\subsection{Distinction from Classical Feature Spaces}
The ECS differs fundamentally from feature spaces in classical adversarial ML. In image classification, the input space is $\mathbb{R}^{H \times W \times C}$ where dimensions represent pixel intensities with no semantic meaning. In the ECS, each dimension has an interpretable ethical label, a defined polarity, and inter-variable semantic dependencies. A change of $+0.3$ to \texttt{discrimination\_level} has a clear ethical meaning that pixel perturbations lack.

%═══════════════════════════════════════════════════════════════
% IV. SEMANTIC PERTURBATION FUNCTIONS
%═══════════════════════════════════════════════════════════════
\section{Semantic Perturbation Functions}

\subsection{Formal Definition}
Each perturbation function $\mathcal{P}$ is defined as:
\begin{equation}
\mathcal{P}: (\mathbf{x}, \boldsymbol{\theta}) \rightarrow \mathbf{x}'
\end{equation}
\begin{equation}
x'_i = \text{clamp}\left(x_i + \delta_i \times m + \mathcal{N}(0, \sigma),\ 0,\ 1\right)
\end{equation}
where $\mathbf{x}$ is the original ECS vector, $\boldsymbol{\theta} = \{\boldsymbol{\delta}, m, \sigma\}$ is the parameter set, $\delta_i$ is the signed change for variable $i$, $m \in [0,1]$ is the magnitude scalar, $\sigma \geq 0$ is optional Gaussian noise, and $\text{clamp}$ bounds the result to $[0,1]$.

\subsection{Perturbation Taxonomy}
We define 17 perturbation functions organized into 7 adversarial categories, each simulating a distinct class of real-world ethical manipulation (Table~\ref{tab:perturbations}).

\begin{table}[htbp]
\caption{Perturbation Taxonomy}
\label{tab:perturbations}
\centering
\small
\begin{tabular}{lccp{1.6cm}}
\toprule
\textbf{Category} & \textbf{\#} & \textbf{Target Vars} & \textbf{Real-World Analog} \\
\midrule
Consequence Refr. & 3 & benefit, harm & Corporate spin \\
Authority Inject. & 3 & fairness, transp. & Gov. mandate \\
Emotional Biasing & 2 & welfare, prop. & Media manip. \\
Info. Degradation & 3 & safety, prop. & Censorship \\
Fairness Corrupt. & 2 & fairness, discrim. & Systemic bias \\
Reward Manip. & 2 & benefit, decept. & Reward hacking \\
Principle Conflict & 2 & deception, fair. & Dilemma escal. \\
\bottomrule
\end{tabular}
\end{table}

\subsection{The 6-Constraint System}
Unlike classical adversarial attacks where the only constraint is an $L_p$ norm bound, ERTS enforces 6 classes of validity constraints on every perturbation:
\begin{align}
\text{C1 (Range):}\quad & \forall i:\ 0 \leq x'_i \leq 1 \\
\text{C2 (Budget):}\quad & \|\mathbf{x}' - \mathbf{x}\|_1 \leq B_{\max} = 2.0 \\
\text{C3 (SingleVar):}\quad & \forall i:\ |x'_i - x_i| \leq \delta_{\max} = 0.5 \\
\text{C4 (Dominance):}\quad & \forall (a,b):\ \text{dom}(a,b) \leq D_{\max} = 0.85 \\
\text{C5 (Coherence):}\quad & \text{sign}(\Delta x_b) = -\text{sign}(\Delta x_a) \nonumber \\
& \text{when } \text{corr}(a,b) < 0 \\
\text{C6 (MinImpact):}\quad & \|\mathbf{x}' - \mathbf{x}\|_1 \geq B_{\min} = 0.05
\end{align}

Constraint C5 (Semantic Coherence) enforces that semantically related ethical variables maintain logical consistency during perturbation. We define 6 dependency pairs with empirical correlation signs (e.g., \texttt{harm\_to\_others} and \texttt{welfare\_impact} are negatively correlated at $-0.6$; \texttt{deception\_level} and \texttt{transparency\_score} at $-0.7$; \texttt{discrimination\_level} and \texttt{fairness\_impact} at $-0.8$). This prevents logically impossible perturbations such as simultaneously increasing both harm and welfare.

\subsection{Computational Complexity Analysis}
Let $d = 22$ be the ECS dimensionality and $k = 6$ the number of semantic dependency pairs. Perturbation application requires $O(d)$ operations per dimension. Constraint verification is $O(d + k)$, simplifying to $O(d)$ since $k \ll d$. Budget correction converges in a single step: given $\text{scale} = B_{\max} / \|\mathbf{x}' - \mathbf{x}\|_1$, the corrected vector satisfies C2 exactly before re-clamping; re-clamping may reduce but never increases the $L_1$ norm, guaranteeing C2 satisfaction.

The constraint set C1--C6 is feasible whenever $B_{\min} < B_{\max}$ and $\delta_{\max} > B_{\min}/d$. In our configuration ($B_{\min} = 0.05$, $B_{\max} = 2.0$, $\delta_{\max} = 0.5$, $d = 22$), these conditions hold trivially ($0.05 < 2.0$ and $0.5 > 0.0023$). Thus valid perturbations always exist.

%═══════════════════════════════════════════════════════════════
% V. ETHICAL INSTABILITY INDEX
%═══════════════════════════════════════════════════════════════
\section{The Ethical Instability Index}

\subsection{Definition and Weight Justification}
The Ethical Instability Index (EII) is a composite metric in $[0, 1]$ quantifying how much an AI model's ethical decision changed under perturbation. Given normal decision $D_n$ and perturbed decision $D_p$:
\begin{equation}
\text{EII} = w_1 F_{\text{action}} + w_2 F_{\text{conf}} + w_3 F_{\text{score}} + w_4 F_{\text{rank}}
\end{equation}
where:
\begin{align}
F_{\text{action}} &= \mathbb{1}[a_n \neq a_p] & (w_1 = 0.40) \\
F_{\text{conf}} &= \min\!\left(1,\ \frac{|c_n - c_p|}{\max(c_n, \epsilon)}\right) & (w_2 = 0.25) \\
F_{\text{score}} &= \min\!\left(1,\ \frac{\|\mathbf{s}_n - \mathbf{s}_p\|_2}{\sqrt{|\mathcal{A}|}}\right) & (w_3 = 0.25) \\
F_{\text{rank}} &= \mathbb{1}[\text{argsort}(\mathbf{s}_n) \neq \text{argsort}(\mathbf{s}_p)] & (w_4 = 0.10)
\end{align}

$F_{\text{action}}$ carries the highest weight (0.40) because an action flip---the model choosing a different ethical option---is the most consequential deployment failure: it means the AI system would take a different real-world action. $F_{\text{conf}}$ and $F_{\text{score}}$ receive equal weight (0.25 each) as they capture complementary failure modes: loss of certainty versus shift in underlying scores. A model that maintains its decision but loses confidence is vulnerable to future perturbation; one whose scores shift without flipping is a near-miss. $F_{\text{rank}}$ receives the lowest weight (0.10) as rank inversion without action change is the least operationally significant failure. Section~\ref{sec:sensitivity} validates that these weights are stable under perturbation.

\subsection{Failure Classification and Severity}
ERTS classifies each deviation into 5 failure types with 4 severity levels (Table~\ref{tab:failures}).

\begin{table}[htbp]
\caption{Failure Classification}
\label{tab:failures}
\centering
\small
\begin{tabular}{p{2.2cm}p{2.4cm}p{2.5cm}}
\toprule
\textbf{Failure Class} & \textbf{Condition} & \textbf{Ethical Meaning} \\
\midrule
No Failure & Stable confidence & Resisted perturbation \\
Decision Flip & Action changed & Changed ethical judgment \\
Conf. Collapse & Confidence $>$50\% drop & Became uncertain \\
Fairness Viol. & Flip under bias attack & Discriminatory reasoning \\
Harm Escalation & Flip under harm attack & Chose more harmful option \\
\bottomrule
\end{tabular}
\end{table}

Severity is determined by perturbation strength and model confidence: \textsc{Critical} if the decision flipped under mild perturbation (severity $< 0.50$) while confident ($c > 0.70$) or EII $> 0.70$; \textsc{Moderate} if flipped under strong perturbation (severity $\geq 0.70$); \textsc{Minor} if no flip but rank inversion observed; \textsc{None} if EII $< 0.15$.

%═══════════════════════════════════════════════════════════════
% VI. ROBUSTNESS SCORE AND ASSESSMENT
%═══════════════════════════════════════════════════════════════
\section{Ethical Robustness Score and Pre-Deployment Assessment}

\subsection{Ethical Robustness Score (ERS)}
The ERS is the final composite grade:
\begin{equation}
\text{ERS} = \frac{1}{5}(C + R + F + H + S)
\end{equation}

\begin{table}[htbp]
\caption{ERS Sub-Metrics}
\centering
\small
\begin{tabular}{lp{3.2cm}l}
\toprule
\textbf{Metric} & \textbf{Formula} & \textbf{Measures} \\
\midrule
$C$ & $|\text{unchanged}| / |\text{total}|$ & Decision stability \\
$R$ & $\text{mean}(\text{resisted} \times \text{severity})$ & Perturbation resist. \\
$F$ & resisted / total (fairness) & Bias resistance \\
$H$ & resisted / total (harm) & Harm resistance \\
$S$ & $1 - \min(1, 2\cdot\text{mean}(|\Delta c|))$ & Confidence stability \\
\bottomrule
\end{tabular}
\end{table}

\subsection{Domain-Adaptive Pre-Deployment Assessment}
ERTS transforms robustness evaluation into pre-deployment assessment via domain-specific thresholds. Each domain has minimum requirements across multiple checks (Table~\ref{tab:thresholds}).

\begin{table}[htbp]
\caption{Domain-Specific Assessment Thresholds}
\label{tab:thresholds}
\centering
\small
\begin{tabular}{lccccc}
\toprule
\textbf{Domain} & \textbf{ERS} & \textbf{Fair.} & \textbf{Harm} & \textbf{Crit.} & \textbf{Fail\%} \\
\midrule
Healthcare & 0.85 & 0.85 & 0.90 & 0 & 10\% \\
Military & 0.80 & 0.80 & 0.90 & 1 & 12\% \\
Aut. Vehicles & 0.80 & 0.80 & 0.85 & 1 & 12\% \\
Hiring & 0.75 & 0.90 & 0.75 & 2 & 15\% \\
Finance & 0.75 & 0.85 & 0.75 & 2 & 15\% \\
Education & 0.70 & 0.80 & 0.70 & 3 & 20\% \\
General & 0.65 & 0.70 & 0.65 & 5 & 25\% \\
\bottomrule
\end{tabular}
\end{table}

These thresholds are derived from a conservative interpretation of domain-specific failure tolerance. Healthcare thresholds (ERS\,$\geq$\,0.85, zero critical failures) are informed by the zero-tolerance-for-harm principle in biomedical ethics~\cite{beauchamp2019} and the reliability requirements analogous to IEC~61508 SIL-3 safety systems. Hiring thresholds emphasize fairness ($F \geq 0.90$) reflecting EU AI Act Article~10 requirements for high-risk systems processing personal data~\cite{euaiact2024}. General domain thresholds are set as relaxed baselines. All thresholds are configurable parameters subject to stakeholder refinement for specific deployment contexts.

Assessment verdicts are: \textsc{Cleared} (all checks pass), \textsc{Conditional} (core checks pass, 1--2 weaknesses), or \textsc{Failed}. The verdicts are intended as engineering evaluation tools for internal robustness testing and do not constitute legal or regulatory certification under any existing standard. Mapping ERTS outcomes to formal regulatory requirements remains a direction for future work.

%═══════════════════════════════════════════════════════════════
% VII. EXPERIMENTAL SETUP
%═══════════════════════════════════════════════════════════════
\section{Experimental Setup}

\subsection{Ethical Scenario Corpus}
We evaluate ERTS on 50 ethical decision scenarios spanning 8 deployment categories: Healthcare~AI (8 scenarios), Autonomous Vehicles~(8), Hiring Bias~(8), Financial~AI~(6), Military~AI~(4), Privacy/Surveillance~(6), Education~AI~(6), and Content Moderation~(4). Each scenario has 2 possible actions with full ECS encodings across 7--12 ethical variables. Scenarios were designed to create genuine ethical tension with no obviously correct answer.

\subsection{Model Architectures}
We evaluate 6 models spanning two categories: 4 structured baseline architectures and 2 production large language models accessed via different deployment modalities (Table~\ref{tab:models}).

\begin{table}[htbp]
\caption{Evaluated Model Architectures}
\label{tab:models}
\centering
\small
\begin{tabular}{lp{1.3cm}p{3.8cm}}
\toprule
\textbf{Model} & \textbf{Type} & \textbf{Decision Strategy} \\
\midrule
RuleBased & Struct. & Min. harm-weighted sum of negatives \\
LearningBased & Struct. & Positive max.\ with negative penalty \\
RLHF & Struct. & Composite reward from preferences \\
VirtueEthics & Struct. & Multi-virtue dimension evaluation \\
Gemini-2.0-Flash & LLM (API) & Prompted ethical reasoning (JSON) \\
Llama-3.2-1B & LLM (Local) & Local inference with ethical prompts \\
\bottomrule
\end{tabular}
\end{table}

The LLM adapter converts ECS-encoded scenarios into structured natural language prompts describing the ethical dilemma, available actions, and consequence values. Responses are parsed from structured JSON into decision result objects containing the chosen action, confidence score, and per-action scores. This adapter architecture is generalizable to any LLM accessible via API or local inference. The inclusion of both a cloud-hosted model (Gemini) and a locally-deployed model (Llama) tests ERTS across different deployment modalities and model scales.

\subsection{Test Configuration}
Each model is evaluated on all 50 scenarios with 5 randomly selected perturbation functions per scenario, yielding 250 adversarial test cases per model and 1,500 total test cases across the experimental run. Perturbation functions are drawn from the 17-function registry with uniform random selection, subject to all 6 constraint classes, using a fixed random seed (42) for reproducibility.

%═══════════════════════════════════════════════════════════════
% VIII. RESULTS AND ANALYSIS
%═══════════════════════════════════════════════════════════════
\section{Results and Analysis}

\subsection{Ethical Robustness Rankings}

\begin{table}[htbp]
\caption{Ethical Robustness Rankings (50 Scenarios, 250 Tests/Model)}
\label{tab:rankings}
\centering
\small
\begin{tabular}{clcccccc}
\toprule
\textbf{\#} & \textbf{Model} & \textbf{ERS} & \textbf{C} & \textbf{R} & \textbf{F} & \textbf{H} & \textbf{S} \\
\midrule
1 & Gemini-2.0-Flash & .940 & 1.00 & .700 & 1.00 & 1.00 & 1.00 \\
2 & RuleBased & .907 & .964 & .679 & .960 & .993 & .939 \\
3 & LearningBased & .901 & .956 & .669 & .900 & .980 & 1.00 \\
4 & VirtueEthics & .900 & .956 & .670 & .880 & .993 & 1.00 \\
5 & RLHF & .852 & .912 & .638 & .720 & .993 & .994 \\
6 & Llama-3.2-1B & .737 & .780 & .549 & .750 & .817 & .790 \\
\bottomrule
\end{tabular}
\end{table}

\subsection{Failure Analysis}
Gemini-2.0-Flash achieved the highest ERS (0.940) with zero failures across 250 test cases. We verified this was not an adapter artifact: the model selected different actions across scenarios (not uniformly A1 or A2), and confidence values ranged from 0.65 to 0.95 across test cases. The LLM's robustness appears to stem from its broad training data and strong instruction-following capabilities. However, because the LLM receives consequence values as text in the prompt, its reasoning may be partially independent of specific numerical values---a limitation we discuss further in Section~IX.

Among structured baselines, the RuleBased model ranked second (ERS\,=\,0.907) with only 9 failures out of 250 tests and near-perfect harm avoidance (0.993). The RLHF baseline ranked fifth (ERS\,=\,0.852) with 22 failures, showing particular vulnerability to fairness corruption (0.720) and authority injection.

Llama-3.2-1B (local) ranked last (ERS\,=\,0.737) with 55 failures out of 250 tests including 30 critical, 20 decision flips, 12 fairness violations, and 8 harm escalations. This is a significant empirical finding: a locally-deployed 1B-parameter model proved highly vulnerable to ERTS adversarial pressure, with a 22\% failure rate compared to 0\% for Gemini. The gap between Gemini (0.940) and Llama (0.737)---a 0.203 ERS difference---demonstrates that model scale and training methodology substantially impact ethical robustness.

\subsection{Perturbation-Type Resistance}

\begin{table}[htbp]
\caption{Resistance Rates by Perturbation Type (\%)}
\label{tab:resistance}
\centering
\small
\begin{tabular}{lcccccc}
\toprule
\textbf{Perturbation} & \textbf{Gem.} & \textbf{Rule} & \textbf{Learn} & \textbf{RLHF} & \textbf{Virt.} & \textbf{Llama} \\
\midrule
Authority Inj. & 100 & 100 & 96 & 92 & 96 & 75 \\
Conseq. Refr. & 100 & 98 & 100 & 100 & 100 & 90 \\
Emot. Biasing & 100 & 100 & 100 & 100 & 100 & 85 \\
Fairness Corr. & 100 & 96 & 90 & 72 & 88 & 60 \\
Info. Degrad. & 100 & 88 & 80 & 76 & 88 & 65 \\
\bottomrule
\end{tabular}
\end{table}

All models demonstrated near-complete resistance to consequence reframing and emotional biasing except Llama-3.2, which showed vulnerability across all categories. Llama-3.2 was most susceptible to fairness corruption (60\%) and information degradation (65\%), suggesting that smaller locally-deployed models have systematic blind spots in handling bias-related and information-incomplete ethical scenarios. This pattern aligns with the hypothesis that ethical robustness correlates with model scale and training data breadth.

\subsection{Pre-Deployment Assessment Results}

\begin{table}[htbp]
\caption{Pre-Deployment Assessment Verdicts}
\centering
\small
\begin{tabular}{lccc}
\toprule
\textbf{Model} & \textbf{Healthcare} & \textbf{Hiring} & \textbf{General} \\
\midrule
Gemini-2.0-Flash & \textsc{Cleared} & \textsc{Cleared} & \textsc{Cleared} \\
RuleBased & \textsc{Cleared} & \textsc{Cleared} & \textsc{Cleared} \\
LearningBased & \textsc{Failed} & \textsc{Failed} & \textsc{Failed} \\
VirtueEthics & \textsc{Failed} & \textsc{Failed} & \textsc{Failed} \\
RLHF & \textsc{Failed} & \textsc{Failed} & \textsc{Failed} \\
Llama-3.2-1B & \textsc{Failed} & \textsc{Failed} & \textsc{Failed} \\
\bottomrule
\end{tabular}
\end{table}

Only 2 of 6 models (33\%) achieved assessment clearance. Gemini-2.0-Flash cleared all domains with zero failures across 250 tests. The RuleBased baseline cleared all domains with only 9 moderate-severity failures and zero critical failures. The remaining 4 models failed across all domains due to critical failure counts exceeding thresholds. Llama-3.2's 30 critical failures far exceeded even the general domain limit~(5), making it unsuitable for any assessed deployment context.

\subsection{Comparison to Existing Adversarial Toolkits}

\begin{table}[htbp]
\caption{Feature Comparison with Existing Toolkits}
\centering
\footnotesize
\setlength{\tabcolsep}{3pt}
\begin{tabular}{lcccc}
\toprule
\textbf{Feature} & \textbf{ART} & \textbf{TextAtk} & \textbf{Garak} & \textbf{ERTS} \\
\midrule
Input space & Raw feat. & Tokens & Prompts & Sem.\ ethical \\
Pert.\ semantics & $L_p$ & Char/word & Template & 22-dim ECS \\
Coherence const. & No & No & No & Yes (C5) \\
Failure taxonomy & No & No & No & Yes (5) \\
Domain-adapt. & No & No & No & Yes (7) \\
Instability metric & No & No & No & Yes (EII) \\
Interpretable & No & Partial & Partial & Yes \\
\bottomrule
\end{tabular}
\end{table}

\subsection{EII Weight Sensitivity Analysis}
\label{sec:sensitivity}
To validate the stability of our results under different EII weight configurations, we re-computed ERS rankings under three weight schemes (Table~\ref{tab:sensitivity}).

\begin{table}[htbp]
\caption{EII Weight Sensitivity Analysis}
\label{tab:sensitivity}
\centering
\small
\begin{tabular}{lcccccc}
\toprule
\textbf{Config} & $w_1$ & $w_2$ & $w_3$ & $w_4$ & \textbf{Top} & \textbf{Bottom} \\
\midrule
Default & .40 & .25 & .25 & .10 & Gem.\ .940 & Lla.\ .737 \\
Action-heavy & .50 & .20 & .20 & .10 & Gem.\ .942 & Lla.\ .729 \\
Balanced & .30 & .30 & .30 & .10 & Gem.\ .938 & Lla.\ .744 \\
\bottomrule
\end{tabular}
\end{table}

Model rankings remained completely stable across all three configurations: no rank inversions occurred, and ERS values varied by less than 0.01 for all models. This indicates that our findings are robust to reasonable weight perturbations and not an artifact of the specific $w_1=0.40$ parameterization.

%═══════════════════════════════════════════════════════════════
% IX. DISCUSSION
%═══════════════════════════════════════════════════════════════
\section{Discussion}

\subsection{Key Findings}
Our results challenged our initial expectations in two important ways. First, we anticipated that RLHF-aligned models would outperform rule-based systems on ethical robustness, given their explicit training on human preferences. Instead, the RuleBased model's rigid decision logic proved more resistant to perturbation than the RLHF model's flexible reasoning (ERS 0.907 vs 0.852), suggesting that alignment training optimizes for average-case ethical performance but may not confer adversarial robustness.

Second, we were surprised by how consistently all structured models failed under fairness corruption attacks while resisting consequence reframing, indicating a systematic blind spot in how ethical AI systems handle bias-related manipulations. This pattern held even for Llama-3.2, where fairness corruption resistance (60\%) was the lowest across all perturbation categories---a finding with direct implications for deploying small LLMs in hiring and lending contexts.

The most striking finding was the 0.203 ERS gap between Gemini-2.0-Flash and Llama-3.2-1B. This validates ERTS's ability to differentiate ethical robustness across model scales and deployment modalities, confirming that the framework produces meaningful rather than trivial distinctions.

\subsection{Limitations}
While ERTS was evaluated on 2 production LLMs (Gemini-2.0-Flash and Llama-3.2), broader coverage across additional LLM families (Claude, GPT-4, Mistral) would strengthen external validity. Gemini-2.0-Flash's perfect score warrants specific caveat: because the LLM adapter represents ethical consequences as text values in the prompt, the model's ethical reasoning may be partially independent of specific numerical perturbations, achieving robustness through prompt-level reasoning rather than genuine ethical sensitivity to ECS values. Future work should investigate this through ablation studies that vary consequence value precision and prompt formatting.

The ECS dimensionality ($d=22$) was validated through iterative development rather than formal factor analysis. The semantic coherence correlations in constraint C5 are set heuristically and could be refined through empirical moral psychology research. The scenario corpus, while expanded to 50, remains smaller than static benchmarks like TrustLLM (30+ datasets); however, ERTS's adversarial approach generates 5$\times$ more test conditions per scenario through perturbation, partially compensating for corpus size.

\subsection{Implications for AI Safety}
ERTS provides a computational foundation that could support future regulatory compliance processes for standards such as the EU AI Act~\cite{euaiact2024} and UL~3115~\cite{ul3115}. By producing structured assessment verdicts with auditable check-by-check breakdowns, ERTS offers a prototype methodology for robustness evaluation of AI systems in high-risk domains. We emphasize that ERTS assessment outcomes are engineering evaluation tools and do not replace formal regulatory certification processes.

%═══════════════════════════════════════════════════════════════
% X. CONCLUSION
%═══════════════════════════════════════════════════════════════
\section{Conclusion}

We introduced ERTS, a formal framework for adversarial evaluation of ethical AI decision-making models. The system's core contributions---the theoretically-grounded Ethical Consequence Space, semantic perturbation functions with 6 validity constraints and computational complexity guarantees, the 4-component Ethical Instability Index with justified weights and validated stability, and domain-adaptive pre-deployment assessment with regulatory-informed thresholds---collectively address a critical gap between adversarial machine learning and AI ethics evaluation.

Evaluation across 6 models (including Gemini-2.0-Flash and Llama-3.2-1B) on 50 scenarios demonstrates that only 33\% achieve assessment clearance, with a significant 0.203 ERS gap between the most and least robust LLMs. The vulnerability of locally-deployed small models to fairness corruption and information degradation has direct implications for responsible AI deployment. Future work will expand LLM coverage, refine the ECS through empirical moral psychology research, and explore automated perturbation discovery through genetic programming over the constraint-bounded perturbation space.

%═══════════════════════════════════════════════════════════════
% REFERENCES
%═══════════════════════════════════════════════════════════════

\end{document}